\documentclass[final]{cvpr}

\usepackage{times}
\usepackage{epsfig}
\usepackage{graphicx}
\usepackage{amsmath}
\usepackage{amssymb}
\usepackage{algorithm}
\usepackage[noend]{algpseudocode}
\usepackage{multirow}
\usepackage{pifont}
\usepackage{xfrac}
\usepackage{diagbox}

\usepackage[pagebackref=true,breaklinks=true,colorlinks,bookmarks=false]{hyperref}

\usepackage{booktabs}
\usepackage[table]{xcolor}
\definecolor{lightgray}{gray}{0.9}
\definecolor{lightblue}{rgb}{0.93,0.95,1.0}
\definecolor{darkgreen}{rgb}{0.0,0.6,0.0}
\definecolor{mypink1}{rgb}{0.858, 0.188, 0.478}
\definecolor{black}{rgb}{0, 0, 0}
\newcommand{\minisection}[1]{\vspace{2mm}\noindent{\textbf{#1}.}}
\newenvironment{tight_itemize}{
\begin{itemize}
  \setlength{\topsep}{0pt}
  \setlength{\itemsep}{2pt}
  \setlength{\parskip}{0pt}
  \setlength{\parsep}{0pt}
}{\end{itemize}}

\usepackage{algorithm}
\usepackage[noend]{algpseudocode}
\usepackage{multirow}
\usepackage{pifont}
\newcommand{\xmark}{\text{\ding{55}}}

\begin{document}

\title{HyperSeg: Patch-wise Hypernetwork for Real-time Semantic Segmentation}

\author{Yuval Nirkin\thanks{Performed this work while an intern at Facebook.}\\
Facebook AI \& Bar-Ilan University\\
\and
Lior Wolf\\
Facebook AI \& Tel Aviv University\\
\and
Tal Hassner\\
Facebook AI}

\maketitle

\begin{abstract}
We present a novel, real-time, semantic segmentation network in which the encoder both encodes and generates the parameters (weights) of the decoder. Furthermore, to allow maximal adaptivity, the weights at each decoder block vary spatially. For this purpose, we design a new type of hypernetwork, composed of a nested U-Net for drawing higher level context features, a multi-headed weight generating module which generates the weights of each block in the decoder immediately before they are consumed, for efficient memory utilization, and a primary network that is composed of novel dynamic patch-wise convolutions. Despite the usage of less-conventional blocks, our architecture obtains real-time performance. In terms of the runtime vs. accuracy trade-off, we surpass state of the art (SotA) results on popular semantic segmentation benchmarks: PASCAL VOC 2012 (val. set) and real-time semantic segmentation on Cityscapes, and CamVid.
The code is available:~\url{https://nirkin.com/hyperseg}.
\end{abstract}

\section{Introduction}
\label{sec:Introduction}

Semantic segmentation plays a crucial role in scene understanding, whether the scene is microscopic, telescopic, captured by a moving vehicle, or viewed through an AR device. New mobile applications go beyond seeking accurate semantic segmentation, and also requiring real-time processing, spurring research into real-time semantic segmentation. This domain has since become a leading test-bed for new architectures and training methods, with the goals of improving both accuracy and speed.
Recent work added capacity~\cite{chen2017rethinking,chen2018encoder} and attention mechanisms~\cite{ho2019axial,tao2020hierarchical,wang2020axial} to improve performance. When runtime is not a concern, the image is often processed multiple times by the model and the results are accumulated. In this paper, we attempt to improve the performance in a different way: by providing the network with additional adaptivity.

\begin{figure}[t]
\centering
\includegraphics[width=1.0\linewidth]{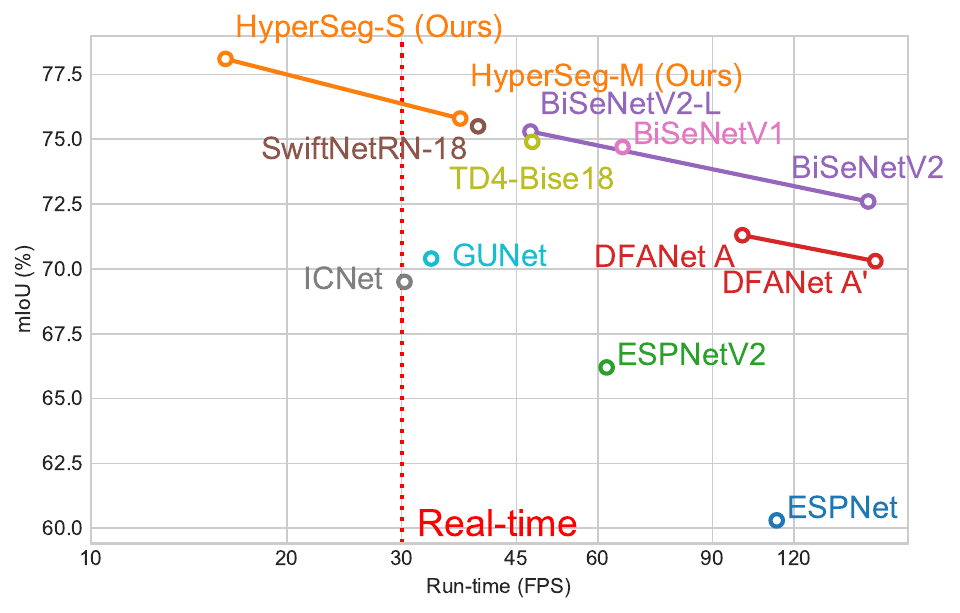}
\caption{\emph{Run-time / accuracy trade-off comparison on the Cityscapes~\cite{cordts2016cityscapes} test set.} Our models (in orange) achieve the best accuracy and the best run-time vs. accuracy trade-off relative to all previous real-time methods.}
\vspace{-4mm}
\label{fig:cityscapes_test_tradeoff}
\end{figure}

We add this adaptivity using a meta-learning technique, often referred to as {\em dynamic networks} or {\em hypernetworks}~\cite{ha2016hypernetworks}. These networks are used for tasks ranging from text analysis~\cite{ha2016hypernetworks,wu2019pay} to 3D modeling~\cite{littwin2019deep,sitzmann2020implicit}, but rarely for generating image-like maps. The reason is that the hypernetworks, as suggested by previous methods, do not fully capture the signals of high resolution images.


Semantic segmentation map are an especially interesting case. They are generated by a coarse to fine pyramid, where each level of the process can benefit from adaptation, since these effects accumulate from one block to the next. Moreover, since every part of the image may contain a different object, such adaptation is best done {\em locally}. 


We thus offer a novel encoder-decoder approach, in which the encoder's backbone is based on recent advances in the field. The encoded signal is mapped to dynamic network weights using an internal U-Net, while the decoder consists of dynamic blocks with spatially varying weights.

The proposed architecture achieves SotA accuracy vs. runtime trade-off on the most widely used benchmarks for this task: PASCAL VOC 2012~\cite{everingham2010pascal}, CityScapes~\cite{cordts2016cityscapes}, and CamVid~\cite{brostow2009semantic}. For CityScapes and CamVid, the SotA accuracy result is obtained under the real-time conditions. Despite using an unconventional architecture that employs locally connected layers with dynamic weights, our method is very efficient (See Fig.~\ref{fig:cityscapes_test_tradeoff} for our run-time / accuracy trade-off relative to other methods.) 

To summarize, our contributions are:
\begin{tight_itemize}
\item A new hypernetwork architecture that employs a U-Net within a U-Net.
\item Novel dynamic patch-wise convolution with weights that vary both per input and per spatial location.
\item SotA accuracy vs. runtime trade-off on the major benchmarks of the field.
\end{tight_itemize}

\section{Related work}
\label{sec:Related work}
\noindent{\bf Hypernetworks.}
Hypernetwork~\cite{ha2016hypernetworks}, are networks that generate weight values for other networks (often referred to as {\em primary networks}). Hypernetworks are useful as a modeling tool, e.g., as implicit functions for image-to-image translation~\cite{de2016dynamic,klocek2019hypernetwork}, 3D scene representation~\cite{littwin2019deep,sitzmann2020implicit}, and also for avoiding compute and data heavy training cycles during neural architecture search (NAS)~\cite{zhang2018graph} and continual learning~\cite{von2019continual}. To our knowledge, however, hypernetworks were never proposed for semantic segmentation, as we propose to do here.

\minisection{Locally connected layers} Connectivity in locally connected layers follows a spatial pattern that is similar to conventional convolutional layers but without weight sharing. Such layers played an important role in the early days of deep learning, mostly due to computational reasons~\cite{dean2012large,raina2009large,uetz2009large}.

Locally connected layers were introduced as an accuracy enhancing component in the context of face recognition, where these were motivated by the need to model each part of the face in a different way~\cite{taigman2014deepface}. However, subsequent face recognition methods tend to use conventional convolutions, e.g.,~\cite{schroff2015facenet}. Partial sharing of weights, in which convolutions are shared within image patches, was proposed for the analysis of facial actions~\cite{zhao2016deep}.

As far as we can ascertain, we are the first to propose locally connected layers in combination with hypernetworks or in the context of semantic segmentation or, more generally, in image-to-image mapping.

\minisection{Semantic segmentation}
Early semantic segmentation methods used feature engineering and often relied on data driven approaches~\cite{hassner2016sifting,hassner2016dense,hassner2012sifts,tau2015dense}.
To our knowledge, Long et al.~\cite{long2015fully} were the first to show end-to-end training of convolutional neural networks (CNN) for semantic segmentation. Their fully convolutional network (FCN) output dense, per-pixel predictions of variable resolutions, based on a classification network backbone. They incorporated skip connections between the early and final layers, for combining coarse and fine information. Subsequent methods added a post processing step based on conditional random fields (CRF) to further refine the segmentation masks~\cite{chen2014semantic,chen2017deeplab,zheng2015conditional}. Nirkin et al. overcame limited, scarce segmentation labels by utilizing motion from videos~\cite{nirkin2018face}. U-Nets~\cite{ronneberger2015u} used encoder-decoder pairs, concatenating the last feature maps of the encoder, in each stride, with corresponding upsampled feature maps from the decoders. 

Some proposed replacing strided convolutions with dilated convolutions, a.k.a. atrous convolutions~\cite{chen2017deeplab,yu2015multi}. This approach produced more detailed segmentations by enlarging the receptive field of the logits but also drastically increased computational costs. Another approach for expanding the receptive field is called spatial pyramid pooling (SPP)~\cite{he2015spatial,zhao2017pyramid}, in which features from different strides are average pooled and concatenated together, after which the information is fused by subsequent convolution layers. Subsequent work combined atrous convolutions with SPP (ASPP), achieving improved accuracy, yet with even higher computational cost~\cite{chen2017deeplab,chen2017rethinking,chen2018encoder}. To further improve accuracy, some proposed inference strategies of applying networks multiple times on multi-scale and horizontally flipped versions of the input image, and combining the results using average pooling~\cite{chen2017rethinking,chen2018encoder}.

Recently, Tao et al.~\cite{tao2020hierarchical} utilized attention to better combine the inference strategy predictions, taking scale into account. Finally, others proposed axial attention, performing attention along the height and width axes separately, to better model long range dependencies~\cite{ho2019axial,wang2020axial} .

\begin{figure*}[t]
\centering
\includegraphics[width=0.95\linewidth]{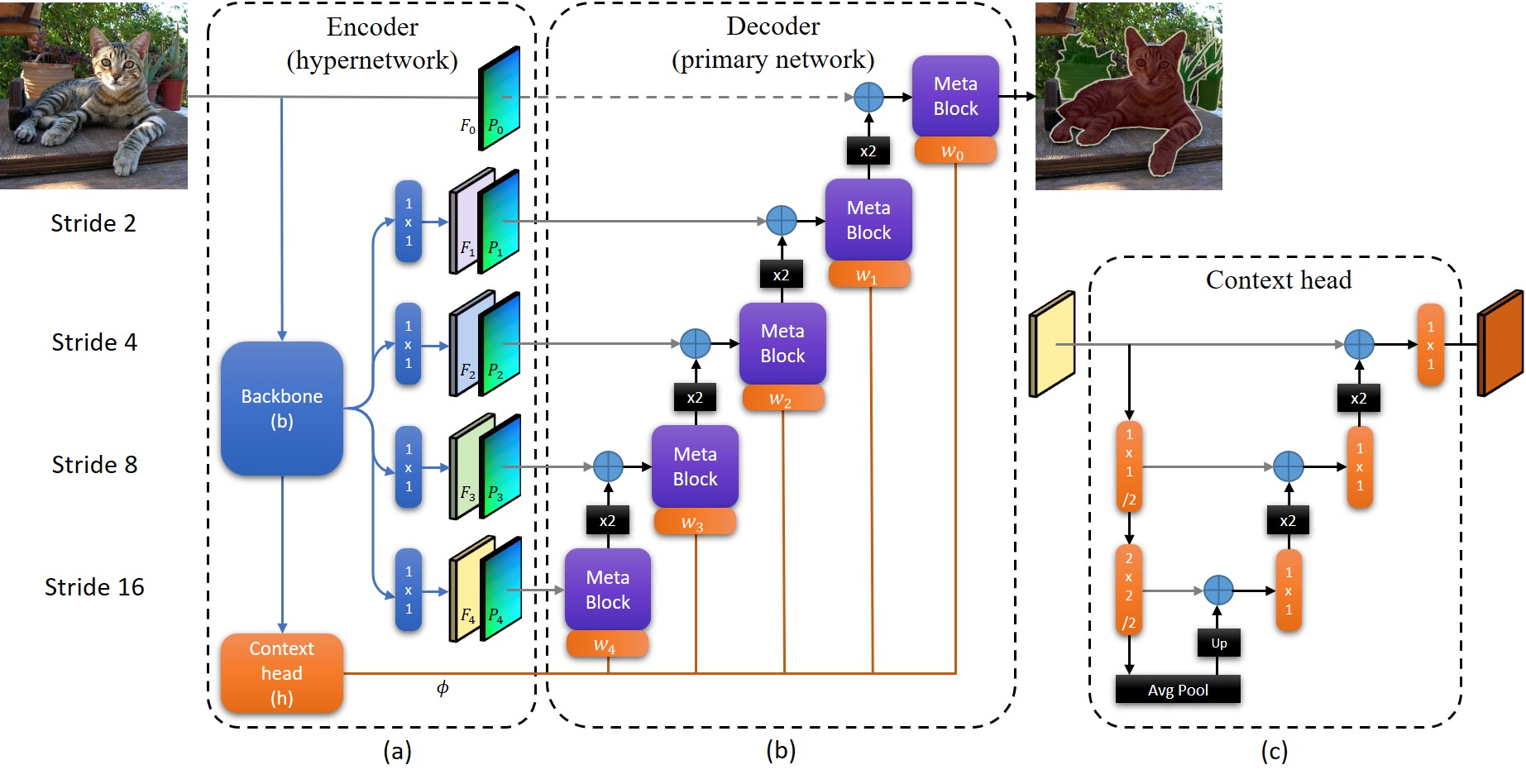}\vspace{-2mm}
\caption{\emph{Method overview.} (a) The hypernetwork encoder based on an EfficientNet~\cite{tan2019efficientnet} backbone, $b$, with its final layer replaced by the context head, $h$. (b) The primary network decoder, $d$, and layers $w_i$ of the weight mapping network, $w$, embedded in each meta block. The input to the decoder, $d$, are the input image and the features, $F_i$, concatenated with positional embedding, $P_i$. Its weights are determined {\em dynamically} for each patch in the image. Gray arrows represent skip connections, $\times$2 blocks are bilinear upsampling, and the blue '+' signs are concatenations. (c) The context head is designed as a nested U-Net. Please see Sec.~\ref{sec:Method overview} for more details.}
\vspace{-2mm}
\label{fig:architecture}
\end{figure*}

\minisection{Real-time segmentation}
The goal of some methods is to achieve the best trade-off between accuracy and computations, with an emphasis on maintaining real-time performance. Real-time methods typically adopt an architecture composed of an encoder based on an efficient backbone and a relatively small decoder.

One early example of inexpensive semantic segmentation is the SegNet~\cite{badrinarayanan2017segnet}, which uses an encoder-decoder architecture with skip connections and transposed convolutions for upsampling. ENet~\cite{paszke2016enet} proposed an architecture based on the ResNet's bottleneck block~\cite{he2016deep}, achieving a high frames per second (FPS) rate but sacrificing considerable accuracy. ICNet~\cite{zhao2018icnet} utilizes fused features extracted from image pyramids, reporting better accuracy than previous methods. GUNet~\cite{mazzini2018guided} performed upsampling guided by the fused feature maps from the encoder, extracted from multi-scale input images. SwiftNet~\cite{orsic2019defense} suggested an encoder-decoder with SPP and $1 \times 1$ convolutions for reducing the number of dimensions before each skip connection.

Subsequent methods benefited from progress in efficient network architecture~\cite{li2019dfanet,yu2020bisenet}, such as depth-wise separable convolutions~\cite{chollet2017xception,howard2017mobilenets} and inverted residual blocks~\cite{sandler2018mobilenetv2} which we also use in our work. BiSeNet~\cite{yu2018bisenet} proposed an additional, coarser downsampling path that is fused with the finer resolution main network before upsampling. BiSeNetV2~\cite{yu2020bisenet} extends BiSeNet by offering a more elaborate fusion of the two branches and additional prediction heads from intermediate layers for boosting the training. Finally, TDNet~\cite{hu2020temporally} proposed a network for video semantic segmentation, by circularly distributing sub-networks over sequential frames, leveraging temporal continuity.

\section{Method}
\begin{figure*}[t]
\centering
\includegraphics[width=0.8\linewidth]{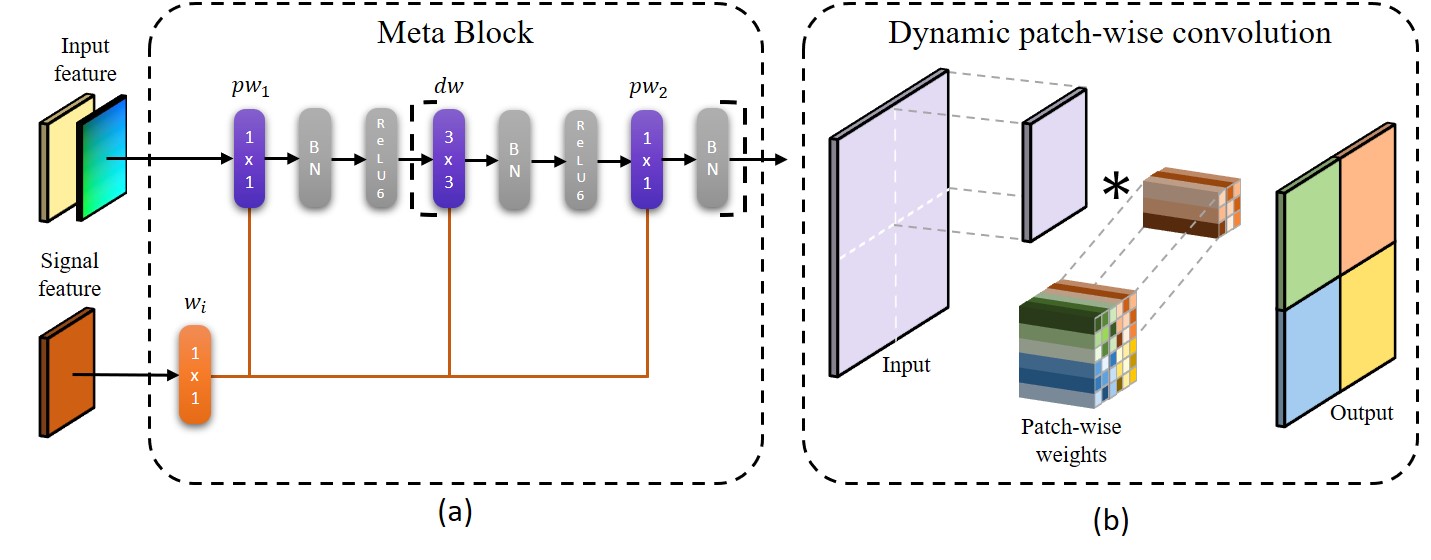}\vspace{-2mm}
\caption{(a) \emph{The meta block based on the inverted residual block}~\cite{sandler2018mobilenetv2}. Each purple layer represents a dynamic patch-wise convolution with weights generated by the orange layer, $w_i$. (b) \emph{Visualization of the dynamic patch-wise convolution operation.} Each color represents weights corresponding to a specific patch and '*' is the convolution operation. Please see Sec.~\ref{subsec:Dynamic patch-wise convolution} for more details.}
\vspace{-2mm}
\label{fig:meta_block}
\end{figure*}

\minisection{Overview}
\label{sec:Method overview}
Our proposed hypernetwork encoder-decoder approach is illustrated in Fig.~\ref{fig:architecture} and Fig.~\ref{fig:meta_block}. Similarly to U-Net--based methods~\cite{ronneberger2015u}, we employ skip connections between corresponding layers of the encoder and the decoder. Our network, however, uses encoder and subsequent blocks which we refer to as the {\em context head} and {\em weight mapper}, in the spirit of hypernetwork design. Skip connections, therefore, connect different encoder levels with the levels of a hierarchical primary network, which serves as our decoder. Moreover, our decoder weights vary between patches at each stride level.

Our proposed model involves three sub-networks: the {\em backbone}, $b$ (shown in blue, in Fig.~\ref{fig:architecture}(a), the {\em context head}, $h$ (orange box in Fig.~\ref{fig:architecture}(a), also detailed in Fig.~\ref{fig:architecture}(c)), and the primary network, acting as the decoder, $d$ (Fig.~\ref{fig:architecture}(b)). In addition, the decoder consists of multiple {\em meta blocks}, visualized in detail in Fig.~\ref{fig:meta_block}(a). Each meta block, $i=0\dots n$, includes an additional weight mapping network component, $w_i$, represented as orange boxes in Fig.~\ref{fig:architecture}(b). 

\minisection{Information flow} The weights of the three networks: $\theta^b$, $\theta^h$, and $\theta^w$, are fixed during inference and learned during the training process, while $\theta^{m_i}$, the weights of the decoder meta block, $m_i$, are predicted dynamically at inference time. 
The encoder's backbone, $b$, maps an input image, $I \in \mathbb{R}^{3 \times H \times W}$, to a set of feature maps, 
$F_i \in \mathbb{R}^{C_i \times \frac{H}{2^i} \times \frac{W}{2^i}},i \in [1,5]$, of different resolutions, where $H$ and $W$ are the number of pixels along the height and width of the image correspondingly.

The context head, 
$h:\mathbb{R}^{C_n \times \frac{H}{2^n} \times \frac{W}{2^n}} \rightarrow \mathbb{R}^{C_n \times \frac{H}{2^n} \times \frac{W}{2^n}}$, maps the last feature map from $b$ to a signal, $\phi$. This signal is then fed to 
$w:\mathbb{R}^{C_n \times \frac{H}{2^n} \times \frac{W}{2^n}} \rightarrow \mathbb{R}^{(\sum_i|\theta^{m_i}|) \times \frac{H}{2^n} \times \frac{W}{2^n}}$ which generates the weights for meta blocks of the primary network, $d$. Note that these weights vary from one spatial location to the next. We define a fixed positional encoding, 
$P^{H,W} \in \mathbb{R}^{2 \times H \times W}$, such that in each position 
$(i,j)$, $P^{H,W}_{i,j}=(\frac{2i - H + 1}{H - 1},\frac{2j - W + 1}{W - 1})$, $i \in [0,H), j \in [0,W)$.

Finally, given the input image and the feature maps, $F_1,\dots,F_n$, their corresponding positional encodings of the same resolutions, $P_0,\dots,P_n$, and the weights, $\theta^d$, the decoder, $d$, outputs the segmentation prediction, $S \in \mathbb{R}^{C \times H \times W}$, where $C$ is the number of classes in the semantic segmentation task.

Our entire network is, therefore, defined by the following set of equations:
\begin{align}
F_1,\dots,F_n &= b(I|\theta^b), \\
\phi &= h(F_n|\theta^h), \\
\theta^{m_i} &= w_i(\phi|\theta^w), i=0,\dots,n,\\
S &= d(\{F_i\},\{P_i\}|\{\theta^{m_i}\})\,,
\end{align}
where the weights of each network are specified explicitly after the separator sign.

\subsection{The encoder and the hypernetwork}
\label{subsec:Signal head}
The first component of the hypernetwork is the backbone network, $b$ (blue box in Fig.~\ref{fig:architecture}(a)). It is based on the EfficientNet~\cite{tan2019efficientnet} family of models. Sec.~\ref{sec:Experimental results} provides details of this network. In our work, the head of the backbone architecture is replaced with our context head, $h$. The backbone outputs the feature map, $F_i$, in each stride. In order to decrease the size of the decoder, we augment it with additional $1 \times 1$ convolutions that reduce the number of channels of each $F_i$ by a factor $r_i$. The exact values for $r_i$ are detailed in Appendix~\ref{sec:Model details}.

The last feature map of $b$, the backbone network, is of size $\frac{H}{2^n} \times \frac{W}{2^n}$. Each pixel in this feature map encodes a patch in the input image. These patches have little overlap, and the limited receptive field can lead to poor results in case of large objects that span multiple patches. The context head, $h$, therefore, combines information from multiple patches.

We detail the structure of $h$ in Fig.~\ref{fig:architecture}(c). Network $h$ uses the nested U-Net structure introduced by Xuebin et al.~\cite{qin2020u2}. In our implementation, we employ $2 \times 2$ convolutions with a stride of two, that output half the number of channels of the input. Such convolutions are computationally cheaper than $3 \times 3$ convolutions, which require padding for the low resolution feature maps processed by $h$, and this padding can significantly increase the spatial resolution. The bottom-most feature map is average pooled to extract the highest level context, and then upsampled to the previous resolution using nearest neighbor interpolation. Finally, in the upsampling path of $h$, at each level, we concatenate the feature map with its corresponding upsampled feature map, followed with a fully connected layer.

While the weight mapping network, $w=[w_0,\dots,w_n]$, is a key part of our hypernetwork, it is more efficient in our hierarchical network to divide $w$ into parts and attach these parts to primary network blocks (Fig.~\ref{fig:architecture}(b)). Hence, instead of directly following $h$, the layers, $w_0,\dots,w_n$, of the weight mapping network are embedded in each of the meta blocks of $d$. The rationale is that the mapping from context to weights incurs a large expansion in memory, which can become a performance bottleneck. Instead, the weights are generated right before they are consumed, minimizing the maximum memory consumption and better utilizing the memory cache. Each $w_i$ is a $1 \times 1$ convolution with channel groups, $g_{w_i}$, and is detailed next.

\subsection{The decoder (the primary network)}
\label{subsec:Decoder}
The decoder, $d$, shown in Fig.~\ref{fig:architecture}(b), consists of $n+1$ meta blocks, $m_0,\dots,m_n$, illustrated in Fig.~\ref{fig:meta_block}(a). Block, $m_0$, corresponds to the input image, and each of the blocks, $m_i$, $i=1..n$, corresponds to the feature map, $F_i$, of the encoder. Each block is followed by bilinear upsampling and concatenation with the next finer resolution feature map.

Unlike conventional schemes, by employing a hypernetwork, the weights of the decoder, $d$, depend on the input image. Moreover, the weights of $d$ are not only conditioned on the input image but also {\em vary between different regions of the image}. With this approach, we can efficiently combine low level information from the stem of the network with high level information from the bottom layers. This allows our method to achieve {\em higher accuracy using a much smaller decoder}, thus enabling real-time performance. The hypernetwork can be seen as a type of attention, and similar to some attention-based methods~\cite{ho2019axial,parmar2019stand}, 
$d$ benefits from knowing the position information of the pixels. For this reason, we augment the input image and the encoder's feature maps with additional positional encoding.

\begin{table}[t]
\begin{center}
{
\begin{tabular}{l c c c c }
  \toprule
  Dataset & Batch & $lr_0$ & $p$ & $t$ \\ [0.5ex] 
  \hline
  PASCAL VOC 2012~\cite{everingham2010pascal} & 32 & $10^{-4}$ & 3.0 & 3.2M \\
  Cityscapes~\cite{cordts2016cityscapes} & 16 & $10^{-3}$ & 0.9 & 1.4M \\
  CamVid~\cite{brostow2009semantic} & 16 & $10^{-3}$ & 2.0 & 240k \\
  \bottomrule
\end{tabular}
}
\end{center}
\caption{\emph{Training hyperparameters for each benchmark.}}
\label{tab:training_details}
\end{table}

\begin{table}[t!]
\begin{center}
\resizebox{1.0\linewidth}{!}{
\begin{tabular}{l c c c c }
  \toprule
  Method & Backbone & mIoU & GFLOPs & Params \\ [0.5ex] 
  \hline
  Auto-DeepLab-L~\cite{liu2019auto} & - & {73.6} & 79.3$^\star$ & 44.42 \\ 
  DeepLabV3~\cite{chen2017rethinking} & ResNet-101 & 78.5 & 249.2$^\star$ & 58.6$^\star$ \\ 
  DFN~\cite{yu2018learning} & ResNet-101 & {79.7} & - & - \\ 
  SDN~\cite{fu2019stacked} & DenseNet-161 & {79.9} & - & 238.5 \\ 
  DeepLabV3+~\cite{chen2018encoder} & Xception-71 & 80.0 & 177 & 43.48 \\  
  HyperSeg-L & EfficientNet-B3 & \textbf{80.6} & \textbf{8.21} & \textbf{39.6} \\
  \bottomrule
\end{tabular}
}
\end{center}
\caption{\emph{Results on the PASCAL VOC 2012, val. set~\cite{everingham2010pascal}.} '$\star$', represents metrics that were computed by us using open source (listed in Appendix~\ref{sec:Open source repositories}).}
\label{tab:pascal_voc_val_quant}
\end{table}

The design of $m_0,\dots,m_n$ is based on the inverted residual block of MobileNetV2~\cite{sandler2018mobilenetv2}: a point-wise convolution, $pw_1$, followed by depth-wise convolution, $dw$, and another point-wise convolution, $pw_2$, without an activation function. Instead of regular convolutions, our network employs dynamic, patch-wise convolutions, described in the next section. For very small patches -- smaller than $4 \times 4$ in our large model; smaller than $8 \times 8$ in our smaller models -- the meta block includes only $pw_1$. The total meta parameters,
$\theta^{m_i} \in \mathbb{R}^{(|\theta^{pw_1}| + |\theta^{dw}| + |\theta^{pw_1}|) \times \frac{H}{2^n} \times \frac{W}{2^n}}$, required by each $m_i$ is the combined meta parameters of all dynamic convolutions in $m_i$:
$\theta^{m_i} = \theta^{pw_1} \cup \theta^{dw} \cup \theta^{pw_2}$. The weights, $\theta^{m_i}$, are generated by the $w_i$ layer, embedded in $m_i$, given the signal, $\phi_i \in \mathbb{R}^{C_{\phi_i} \times \frac{H}{2^n} \times \frac{W}{2^n}}$. At inference, the batch-normalization layers of $m_i$ are fused with $w_i$; more details are provided in Appendix~\ref{sec:Batch normalization fusion}.

Employing the full signal in each $m_i$ is inefficient both computationally and in the number of trainable parameters, because $\phi$ is directly mapped into a large number of weights. We thus instead divide the channels of $\phi$ into parts, $C_{\phi_0},\dots,C_{\phi_n}$, which are relative in size to the required number of weights of each meta block. The division of the channels is defined using the following procedure:
\begin{multline}
C_{\phi_0},\dots,C_{\phi_n} = \text{divide\_channels}( \\
C_n,max(g_{w_0},\dots,g_{w_n}),|\theta^{m_0}|,\dots,|\theta^{m_n}|),
\end{multline}
where divide\_channels$(\cdot)$ is detailed in Appendix~\ref{sec:Feature division algorithm}. This routine ensures that each part is proportional to its allocated signal channels, is divisible by $max(g_{w_0},\dots,g_{w_n})$ for the grouped convolutions in $w$, and is allocated a minimal number of channels.

The number of groups, $g_{w_i}$, is an important hyperparameter, since it controls the amount of computations and trainable parameters invested in producing the weights for $m_i$. Increasing $g_{w_i}$ reduces the computations and trainable parameters in direct proportions, as can be seen from the following equations:
\begin{align}
    |\theta^{w_i}| &= \frac{|\theta^{m_i}| \cdot C_{\phi_i}}{g_{w_i}}, \label{eq:params} \\
    \text{FLOPs}_{w_i} &= \frac{|\theta^{m_i}| \cdot C_{\phi_i} \cdot \frac{H}{2^n} \cdot \frac{W}{2^n}}{g_{w_i}}. \label{eq:flops}
\end{align}
The effect of different values of $g_{w_i}$ is studied in Sec.~\ref{subsec:Cityscapes dataset}.
The exact values of $g_{w_i}$ used in our tests are reported in Appendix~\ref{sec:Model details}.

\begin{table*}[t!]
\begin{center}
{\begin{tabular}{l c c c c c c c}
\toprule
\multirow{2}{*}{Method} & \multirow{2}{*}{Backbone} & \multirow{2}{*}{Resolution} & \multicolumn{2}{c}{mIoU (\%)}
& \multirow{2}{*}{FPS} & \multirow{2}{*}{GFLOPs} & Params  \\ [0.5ex] 
  & & & val & test & & & (M) \\
  \hline
  ERFNet~\cite{romera2017erfnet} & - & $1024 \times 512$ & - & 69.7 & 41.7 & 21.7$^\star$ & 2.0$^\star$ \\
  ESPNet~\cite{mehta2018espnet} & ESPNet & $1024 \times 512$ & - & 60.3 & 112.9 & - & - \\
  ESPNetV2~\cite{mehta2018espnet} & ESPNetV2 & $1024 \times 512$ & 66.4 & 66.2 & 61.9$^\star$ & 2.7 & \textbf{1.3}$^\star$ \\
  ICNet~\cite{zhao2018icnet} & PSPNet50 & $2048 \times 1024$ & - & 69.5 & 30.3 & - & -  \\
  GUNet~\cite{mazzini2018guided} & DRN-D-22 & $1024 \times 512$ & 69.6 & 70.4 & 33.3 & - & - \\
  DFANet A'~\cite{li2019dfanet} & Xception A & $1024 \times 512$ & - & 70.3 & \textbf{160.0} & \textbf{1.7} & 7.8 \\
  DFANet A~\cite{li2019dfanet} & Xception A & $1024 \times 1024$ & - & 71.3 & 100.0 & 3.4 & 7.8 \\
  SwiftNetRN-18~\cite{orsic2019defense} & ResNet18 & $2048 \times 1024$ & 75.4 & 75.5 & 39.9 & 104.0 & 11.8 \\
  BiSeNetV1~\cite{yu2018bisenet} & ResNet18 & $1536 \times 768$ & 74.8 & 74.7 & 65.5 & 75.2$^\star$ & 49.0 \\
  BiSeNetV2~\cite{yu2020bisenet} & - & $1024 \times 512$ & 73.4 & 72.6 & 156.0 & 21.2 & - \\ 
  BiSeNetV2-L~\cite{yu2020bisenet} & - & $1024 \times 512$ & 75.8$^1$ & 75.3 & 47.3 & 118.5 & - \\ 
  TD4-Bise18~\cite{hu2020temporally} & BiseNet18 & $2048 \times 1024$ & 75.0 & 74.9 & 47.6 & - & - \\
\midrule
HyperSeg-M & EfficientNet-B1 & $1024 \times 512$ & 76.2 & 75.8 & 36.9 & 7.5 & 10.1 \\ 
  HyperSeg-S & EfficientNet-B1 & $1536 \times 768$ & \textbf{78.2} & \textbf{78.1} & 16.1 & 17.0 & 10.2 \\

  \bottomrule
\end{tabular}
}
\end{center}
\caption{\emph{Real-time semantic segmentation results on Cityscapes~\cite{cordts2016cityscapes}.} '-' Implies that the metric was not reported. '$\star$', denotes that the specific metric was computed by us using available open source (listed in Appendix~\ref{sec:Open source repositories}). $^1$Reported using horizontal mirroring and multi-scale (confirmed by open source).
\vspace{-2mm}
}
\label{tab:cityscapes_quant}
\end{table*}

\subsection{Dynamic patch-wise convolution}
\label{subsec:Dynamic patch-wise convolution}
We illustrate the operation of the dynamic patch-wise convolution (DPWConv), the layers, $pw_1$, $dw$, and $pw_2$ of $m_i$, in Fig.~\ref{fig:meta_block}(b). Given an input feature map, $X\in\mathbb{R}^{C_{in} \times H \times W}$, and a grid of weights, 
$\theta\in\mathbb{R}^{C_{out} \times \frac{C_{in}}{G} \times K_h \times K_w \times N_h \times N_w}$, where $C_{in}$ and $C_{out}$ are the channel numbers for the input and output, $G$ is the number of channel groups, $H$ and $W$ are the input's height and width, $K_h$ and $K_w$ are the height and width of the kernel, and $N_h$ and $N_w$ are the number of patches along the height and width axes, we define output patches as follows:
\begin{equation}\label{eq:Oij}
O_{i,j}=X_{i,j}*\theta_{i,j},
\end{equation}
where $*$ is the convolution operation, $i \in [0, N_h)$ and $j \in [0, N_w)$ are the patch indices, $X_{i,j}$ is a patch of $X$ in the grid location $(i,j)$, and $\theta_{i,j}$ are the corresponding weights from the weights grid. We first apply padding to the entire input feature map, $X$, and then at each patch, $X_{i,j}$, we wrap the adjacent pixels from neighboring patches.

\section{Experimental results}
\label{sec:Experimental results}
We experiment on three popular benchmarks: PASCAL VOC 2012~\cite{everingham2010pascal}, Cityscapes~\cite{cordts2016cityscapes}, and CamVid~\cite{brostow2009semantic}. We report results using the following standard measures: class mean intersection over union (mIoU), frames per second (FPS), billion floating point operations (GFLOP), and number of trainable parameters. 

FPS is measured using established protocols~\cite{orsic2019defense}: We record FPS for the elapsed time between data upload to GPU through to prediction download. Our model is implemented in PyTorch without specific optimizations. Finally, we use a batch size of $1$ to simulate real-time inference. 
Similar to most previous methods, we measure FPS on a NVIDIA GeForce GTX 1080TI GPU (i7-5820k CPU and 32GB DDR4 RAM). GFLOPs and trainable parameters are calculated using the pytorch-OpCounter library~\cite{pytorchOpCounter}, also used by others~\cite{orsic2019defense}.

We experiment with large, medium, and small variants of our model, {\em HyperSeg-L}, {\em HyperSeg-M}, and {\em HyperSeg-S}, respectively. The models share the same template and are named according to their size as reflected by their parameter numbers. Both HyperSeg-M and HyperSeg-S omit the finest resolution level of $d$; we bilinearly upsample predictions to the input resolution from their previous level. In HyperSeg-S the channels of the layers in $m_i$ are halved, relative to those of the largest model. We provide model backbone and resolution details, separately, for each experiment. For other hyperparameter values, see Appendix~\ref{sec:Model details}.

\begin{figure}[t]
\centering
\includegraphics[clip,trim=0mm 3mm 0mm 0mm, width=0.96\linewidth]{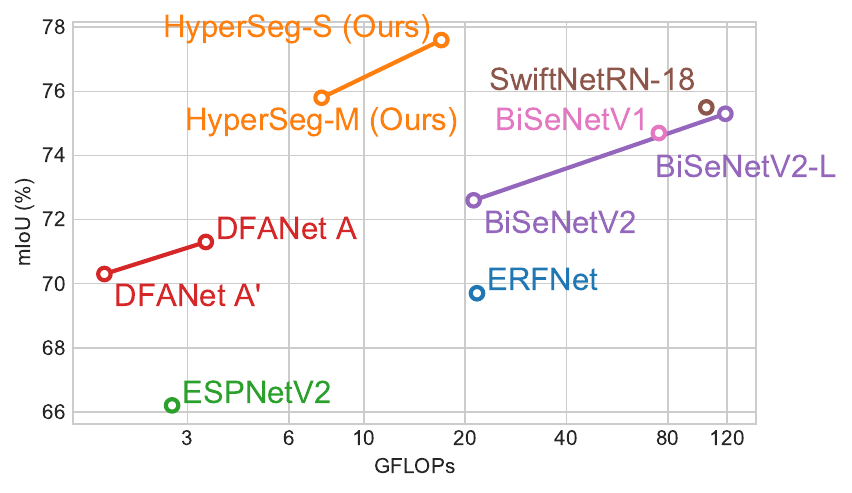}\vspace{0mm}
\caption{
\emph{GFLOPs vs. accuracy trade-off on Cityscapes~\cite{cordts2016cityscapes}. Our models (in orange) attain a significantly better trade-off than previous methods.}
\vspace{-4mm}
}
\label{fig:cityscapes_test_gflops_tradeoff}
\end{figure}

\begin{figure*}[t]
\centering
\includegraphics[clip,trim=0mm 3mm 0mm 0mm, width=0.94\textwidth]{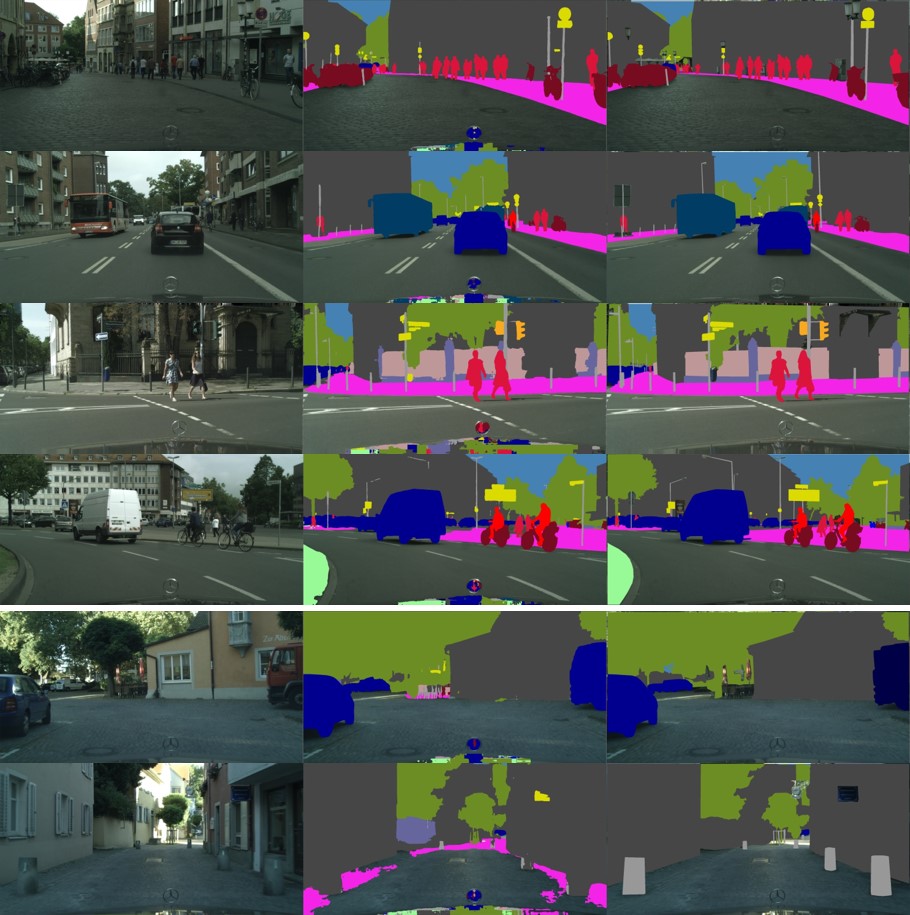}\vspace{0mm}
\caption{
\emph{Qualitative results on Cityscapes~\cite{cordts2016cityscapes} validation set images.} Left to right: input, our result, and ground truth. The first four rows showcase our model's performance in diverse scenes. The last two rows provide sample failures. Please note that the reflective car hood region is ignored in evaluation. \vspace{-3mm}
}
\label{fig:cityscapes_val_qualitative}
\end{figure*}

\subsection{Training details}
\label{subsec:Training details}
We initialize our network using weights pretrained on ImageNet~\cite{russakovsky2015imagenet} for $\theta^b$, and using random values sampled from the normal distribution for $\theta^h$ and $\theta^w$. The Adam optimizer~\cite{kingma2014adam} was used for training, with $\beta_1=0.5$ and $\beta_2=0.999$. Following others~\cite{chen2017rethinking,chen2018encoder}, we use a polynomial learning rate scheduling that decays the initial learning rate, $lr_0$, after $i$ iterations by a factor of $(1 - \frac{i}{t})^{p}$, where $t$ is the total number of iterations and $p$ is a scalar constant. The exact values for each dataset are listed in Tab.~\ref{tab:training_details}.

For the Cityscapes and Camvid benchmarks, we apply the following image augmentations: random resize with scale range $[0.5,2.0]$, crop, and horizontal flipping with probability $0.5$. For PASCAL VOC we use a similar horizontal flip, and random resize with scale range of $[0.25,1.0]$. We further randomly rotate the images, in the range of $-30^{\circ}$ to $30^{\circ}$, jitter colors to manipulate brightness, contrast, saturation, and hue, and finally pad the images to a resolution of $512\times 512$. We train all our models on two Volta V100 32GB GPUs.

\subsection{PASCAL VOC 2012 benchmark tests}
\label{subsec:PASCAL VOC 2012 dataset}
PASCAL VOC 2012~\cite{everingham2010pascal} contains images of varying resolutions, up to $500\times 500$, representing 21 classes (including a class for the background). This set originally contained 1,464 training, 1,449 validation, and 1,456 test images. Its training set was later extended by others to a total of 10,582 images~\cite{hariharan2011semantic}. This set is not typically used to evaluate real-time segmentation methods but the low resolution images allow for quick experimentation. For this reason, we chose this benchmark for our initial tests.

Tab.~\ref{tab:pascal_voc_val_quant} reports accuracy, FLOPs, and number of trainable parameters for our model and those of existing work. We chose methods that reported results on the PASCAL VOC validation set, without inference strategies (e.g., without horizontal mirroring and multi-scale testing). Besides increasing inference times by several factors, these techniques can blur the contributions of the underlying methods. As evident from the results, compared with previous work, our methods achieve the best mIoU, with lower GFLOPs and a smaller number of trainable parameters.

\subsection{Cityscapes benchmark tests}
\label{subsec:Cityscapes dataset}
Cityscapes~\cite{cordts2016cityscapes} provides 5k images of urban street scenes, labeled with 19 classes. Image resolutions are $2048\times1024$ pixels, and are typically downsampled or cropped during training time. The images are partitioned into 2,975 training, 500 validation, and 1,525 test images.

Tab.~\ref{tab:cityscapes_quant} compares variants of our approach using the EfficientNet-B1 backbone~\cite{tan2019efficientnet} operating on different resolutions, with previous methods. We only show previous work considered to be {\em fast}: Methods that run at 10 FPS or faster and report test set mIoU. Our models achieve the best accuracy on both validation and test sets, as well as the best trade-off between accuracy and run-time performance. The trade-off comparison can be best seen in Fig.~\ref{fig:cityscapes_test_tradeoff}.

Importantly, our model incurs a large run-time penalty, due to the unoptimized operations of DPWConv. Fig.~\ref{fig:cityscapes_test_gflops_tradeoff} shows the trade-off between GFLOPs and accuracy of our method relative to previous work. Evidently, our method achieves a significantly better trade-off than other methods. While GFLOPs does not directly correlate with FPS, it does suggest the potential run-time performance, once all our functions are optimized.

Fig.~\ref{fig:cityscapes_val_qualitative} provides qualitative results of our HyperSeg-S model on Cityscapes validation images. Our model produces high quality segmentations without any apparent artifacts, due to the partitioning of images into patches. The last two rows of Fig.~\ref{fig:cityscapes_val_qualitative} offer sample failure cases. In the second row from the bottom, our model confuses a truck with a car. In the last row, our model fails to segment the poles and mistakenly labels pixels as wall or sidewalk.

\subsection{CamVid benchmark tests}
\label{subsec:CamVid dataset}
CamVid offers 701 images of driving scenes, similar to those of Cityscapes, labeled for 11 classes~\cite{brostow2009semantic}. All images share the same resolution, $960\times 720$. The images are partitioned into 367 training, 101 validation, and 233 test images. Following the training protocol used by all of our baselines, we train on both the training and validation sets. 

Tab.~\ref{tab:camvid_quant} compares our approach with previous real-time methods pretrained on ImageNet~\cite{russakovsky2015imagenet}. For a fair comparison, we exclude methods that use additional outside data, other than ImageNet and CamVid. We test two variations of our model, both using the EfficientNet-B1 backbone~\cite{tan2019efficientnet}. HyperSeg-S operates on resolutions of $768\times 576$ and HyperSeg-L on $1024\times 768$. Both models achieve SotA mIoU by a margin relative to that reported by the previous SotA, with HyperSeg-S running at 38 FPS.

Even without outside data, our method outperforms SotA results reported by methods that use Cityscapes as additional training data: The best results using Cityscapes was reported by BiSeNetV2-L~\cite{yu2020bisenet}, which improves its performance from 73.2\% mIoU, when trained without additional data, to 78.5\% with this data. This is still lower, by a margin, than our method: 79.1\% with 16.6FPS. In fact, their result is almost identical to our 38FPS network. Both variants of our method do not use any additional training data.

\minisection{Ablation Study}
We performed ablation studies on the CamVid dataset~\cite{brostow2009semantic}, to show the contribution of our meta-learning approach and the effect of using different backbones. The results are reported in Tab.~\ref{tab:camvid_quant}.

In the first six middle experiments, for each model configuration we replace the EfficientNet-B1 backbone with different backbones: ResNet18, PSPNet18, and PSPNet50. We have explicitly chosen backbones that were used by previous methods. In our implementation, a fully connected layer transforms the last feature map before it is fed to the context head, to 1280 channels for the ResNet18 and PSPNet18 backbones, and 2048 channels for the PSPNet50 backbone. In the PSPNet backbones we do not use dilations in any of the convolutions. In the experiments marked with ``w/o DPWConv'', we replace all the dynamic patch-wise convolutions with regular convolutions, effectively eliminating all the meta-learning elements of our method.

The results clearly show that the EfficientNet-B1 backbone is superior to the ResNet18, PSPNet18, and PSPNet50 backbones, yet {\em our method still outperforms previous methods with those backbones}. Finally, removing meta-learning from our method causes a 1.1\% drop in accuracy for the HyperSeg-S configuration, and a reduction of 0.7\% in accuracy for the HyperSeg-L configuration, with only a slight improvement in FPS, {\em showing that meta-learning is an integral part of our method}.

\begin{table}[t!]
\centering{
\resizebox{1.0\linewidth}{!}{
{\begin{tabular}{l c c c c c c}
\toprule
  \multirow{2}{*}{Method} & \multirow{2}{*}{Backbone} & mIoU & \multirow{2}{*}{FPS} & Params   \\ [0.5ex] 
  & & (\%) & & (M) \\
  \hline
  ICNet~\cite{zhao2018icnet} & PSPNet50 & 67.1 & 27.8 & - \\
  DFANet A~\cite{li2019dfanet} & Xception A & 64.7 & 120.0 & \textbf{7.8} \\
  SwiftNetRN-18~\cite{orsic2019defense} & ResNet18 & 72.6 & 85.8$^\star$ & 11.8 \\
  BiSeNetV1~\cite{yu2018bisenet} & ResNet18 & 68.7 & 116.2 & 49.0 \\
  BiSeNetV2~\cite{yu2020bisenet} & - & 72.4 & \textbf{124.5} & - \\ 
  BiSeNetV2-L~\cite{yu2020bisenet} & - & 73.2 & 32.7 & - \\ 
  TD4-PSP18~\cite{hu2020temporally} & PSPNet18 & 72.6 & 25.0 & - \\
  TD2-PSP50~\cite{hu2020temporally} & PSPNet50 & 76.0 & 11.1 & - \\
  \hline
  HyperSeg-S & ResNet18 & 77.0 & 32.5 & 16.2 \\
  HyperSeg-L & ResNet18 & 77.1 & 11.5 & 16.7 \\ 
  HyperSeg-S & PSPNet18 & 76.6 & 31.3 & 17.2 \\
  HyperSeg-L & PSPNet18 & 77.5 & 11.4 & 17.6 \\ 
  HyperSeg-S & PSPNet50 & 77.1 & 9.3 & 57.6 \\
  HyperSeg-L & PSPNet50 & 77.9 & 2.2 & 67.9 \\ 
  HyperSeg-S w/o DPWConv & EfficientNet-B1 & 77.3 & 45.5 & 9.9 \\
  HyperSeg-L w/o DPWConv  & EfficientNet-B1 & 78.4 & 21.6 & 10.3 \\ 
  \hline
  HyperSeg-S & EfficientNet-B1 & 78.4          & 38.0 & 9.9 \\
  HyperSeg-L & EfficientNet-B1 & \textbf{79.1} & 16.6 & 10.2 \\ 
  \bottomrule
\end{tabular}
}
}
}
\caption{\emph{Real-time semantic segmentation results on CamVid~\cite{brostow2009semantic}} (test set; no outside data).
Top: previous methods. Middle: ablation study. The first six rows are variants of our models with different backbones. In comparison to the baselines, we improve accuracy and increase runtime. In the ``w/o DPWConv'' variants, we replace the dynamic depth-wise convolutions with regular convolutions. {Those variants achieve lower accuracy compared to our full method.} Bottom: Our full method.
$^\star$Computed by us using open source.
\vspace{-2mm}}
\label{tab:camvid_quant}
\end{table}

\section{Conclusions}
We propose to marry autoencoders with hypernetworks for the task of semantic segmentation. In our scheme, the hypernetowrk is a composition of three networks: the backbone of the semantic segmentation encoder, $b$, a context head, $h$, in the form of an internal U-Net, and multiple weight mapping heads, $w_i$. The decoder is a multi-block decoder, where each block, $d_i$, implements locally connected layers. The outcome is a new type of U-Net that is able to dynamically and locally adapt to the input, thus holding the potential to better tailor the segmentation process to the input image. As our experiments show, our method outperforms the SotA methods, in this very competitive field, across multiple benchmarks.

{\small
\bibliographystyle{ieee_fullname}
\bibliography{camera_ready_arxiv}
}

\appendix

\section{Feature division algorithm}
\label{sec:Feature division algorithm}
Algorithm.~\ref{alg:divide_channels} describes the signal channel division algorithm referred to in Sec.~\ref{subsec:Decoder}. The algorithm starts by dividing the channels, $C$, into units of size $s_u$. In our experiments $s_u = max(g_{w_0},\dots,g_{w_n})$, which assures that the divided channels will be divisible by their corresponding $g_{w_i}$, which are all power of 2.

Each weight is first allocated a single unit in order to make sure minimal channel allocation for each of the weights. The units are divided by their total number relative to the sum of the weights, starting from the large weights. This gives priority to the smaller weights, which receive the remainder of the allocations that were rounded down.

\begin{algorithm*}[!t]
\caption{Divides the channels, $C$, in unit size, $s_u$, into chunks relative to the weights, $w_0,\dots,w_n$.}
\label{alg:divide_channels}
\begin{algorithmic}[1]

\Procedure{divide\_channels}{$C,s_u,w_0,\dots,w_n$}
    \State $total\_units \leftarrow \frac{C}{s_u}$
    \State $w \leftarrow sort(w_0,\dots,w_n)$   \Comment{Descending order}
    \State $r \leftarrow \frac{total\_units}{\sum_{i=0}^{n}w_i}$  \Comment{Units to weights ratio}
    \State $out \leftarrow \{ s_u | \text{ for each } w_i \in w\}$ \Comment{Each weight group should be allocated with at least one unit}
    \State $total\_units \leftarrow total\_units - |out|$
    \State $i \leftarrow 0$
    \While{$total\_units \not= 0$}
         \If{$i = n \textbf{ or } \lfloor w_i \cdot r \rfloor \leq 1$}
            \State $curr\_units \leftarrow total\_units$
        \Else
        \State $curr\_units \leftarrow \lfloor w_i \cdot r \rfloor - 1$
        \EndIf
        \State $out_i \leftarrow out_i + curr\_units \cdot s_u$
        \State $total\_units \leftarrow total\_units - curr\_units$
        \State  $i \leftarrow i + 1$
    \EndWhile
    \State
    \Return $out$
\EndProcedure

\end{algorithmic}
\end{algorithm*}

\section{Model details}
\label{sec:Model details}
The hyperparameters of each of our five models are laid out in Tab.~\ref{tab:model_details}. Variables $r_i$ are the reduction factors, corresponding to each of the features maps, $F_i$, from $b$, defined in Sec.~\ref{subsec:Signal head}. $g_{w_i}$, also defined in Sec.~\ref{subsec:Signal head}, equal 16 across all levels for our earliest model, HyperSeg-L (PASCAL VOC), and their values were experimentally adapted for the newer models trained on CamVid and Cityscapes.

The last rows detail the number of feature map channels in each $m_i$. A single arrow, $C_{in} \rightarrow C_{out}$, denotes a single $1 \times 1$ convolution $pw_1:\mathbb{R}^{C_{in} \times \frac{H}{2^i} \times \frac{W}{2^i}} \rightarrow \mathbb{R}^{C_{out} \times \frac{H}{2^i} \times \frac{W}{2^i}}$, and two arrows, $C_{in} \rightarrow C_{hidden} \rightarrow C_{out}$, specify the channels of the full meta block (described in Sec.~\ref{subsec:Decoder}): 
\begin{align}
pw_1&: \mathbb{R}^{C_{in} \times \frac{H}{2^i} \times \frac{W}{2^i}} \rightarrow \mathbb{R}^{C_{hidden} \times \frac{H}{2^i} \times \frac{W}{2^i}}, \\
dw&: \mathbb{R}^{C_{hidden} \times \frac{H}{2^i} \times \frac{W}{2^i}} \rightarrow \mathbb{R}^{C_{hidden} \times \frac{H}{2^i} \times \frac{W}{2^i}}, \\
pw_2&: \mathbb{R}^{C_{hidden} \times \frac{H}{2^i} \times \frac{W}{2^i}} \rightarrow \mathbb{R}^{C_{out} \times \frac{H}{2^i} \times \frac{W}{2^i}}
\end{align}

Our PASCAL VOC model was trained using the cross entropy loss; all other models were trained using bootstrapped cross entropy loss~\cite{reed2014training}.

\begin{table*}[t!]
\centering{
\begin{tabular}{@{}lccccc@{}}
\toprule
\multirow{2}{*}{Params} & \multirow{2}{*}{\begin{tabular}[c]{@{}c@{}}HyperSeg-L\\ (PASCAL VOC)\end{tabular}} & \multirow{2}{*}{\begin{tabular}[c]{@{}c@{}}HyperSeg-S\\ (Cityscapes)\end{tabular}} & \multirow{2}{*}{\begin{tabular}[c]{@{}c@{}}HyperSeg-M\\ (Cityscapes)\end{tabular}} & \multirow{2}{*}{\begin{tabular}[c]{@{}c@{}}HyperSeg-S\\ (CamVid)\end{tabular}} & \multirow{2}{*}{\begin{tabular}[c]{@{}c@{}}HyperSeg-L\\ (CamVid)\end{tabular}} \\
  & & & & & \\ \midrule
Backbone            & EfficientNet-B3 & EfficientNet-B1 & EfficientNet-B1 & EfficientNet-B1 & EfficientNet-B1 \\
Resolution          & $512 \times 512$ & $1536 \times 768$ & $1024 \times 512$ & $768\times 576$ & $1024\times 768$ \\
$r_1,\dots,r_5$     & $\sfrac{1}{4}$, $\sfrac{1}{4}$, $\sfrac{1}{4}$, $\sfrac{1}{4}$, $\sfrac{1}{4}$ &
                      -, $\sfrac{2}{5}$, $\sfrac{1}{4}$, $\sfrac{1}{5}$, $\sfrac{1}{6}$              &
                      $\sfrac{1}{4}$, $\sfrac{1}{4}$, $\sfrac{1}{4}$, $\sfrac{1}{4}$, $\sfrac{1}{4}$ &
                      $\sfrac{1}{4}$, $\sfrac{1}{4}$, $\sfrac{1}{4}$, $\sfrac{1}{4}$, $\sfrac{1}{4}$ &
                      $\sfrac{1}{4}$, $\sfrac{1}{4}$, $\sfrac{1}{4}$, $\sfrac{1}{4}$, $\sfrac{1}{4}$ \\
                      
$g_{w_0},\dots,g_{w_5}$     & 16, 16, 16, 16, 16, 16 & 
                              -, 4, 16, 8, 16, 32    &
                              -, 4, 16, 8, 16, 32    &
                              -, 8, 16, 32, 32, 64   &
                              8, 8, 16, 32, 32, 64   \\
                      
$m_5$ channels      & $98 \xrightarrow{} 96$ &
                      $130 \xrightarrow{} 32$ &
                      $82 \xrightarrow{} 64$ &
                      $82 \xrightarrow{} 64$ &
                      $82 \xrightarrow{} 64$ \\

$m_4$ channels      & $132 \xrightarrow{} 34$ &
                      $62 \xrightarrow{} 16$ &
                      $94 \xrightarrow{} 32$ &
                      $94 \xrightarrow{} 32$ &
                      $94 \xrightarrow{} 32$ \\
                      
$m_3$ channels      & $48 \xrightarrow{} 96 \xrightarrow{} 12$ &
                      $26 \xrightarrow{} 8$ &
                      $44 \xrightarrow{} 16$ &
                      $44 \xrightarrow{} 16$ &
                      $44 \xrightarrow{} 16$ \\

$m_2$ channels      & $22 \xrightarrow{} 44 \xrightarrow{} 8$ &
                      $14 \xrightarrow{} 28 \xrightarrow{} 8$ &
                      $24 \xrightarrow{} 48 \xrightarrow{} 16$ &
                      $24 \xrightarrow{} 48 \xrightarrow{} 16$ &
                      $24 \xrightarrow{} 48 \xrightarrow{} 16$ \\
                      
$m_1$ channels      & $16 \xrightarrow{} 32 \xrightarrow{} 6$ &
                      $26 \xrightarrow{} 52 \xrightarrow{} 19$ &
                      $34 \xrightarrow{} 68 \xrightarrow{} 19$ &
                      $22 \xrightarrow{} 44 \xrightarrow{} 12$ &
                      $22 \xrightarrow{} 44 \xrightarrow{} 16$ \\
                      
$m_0$ channels      & $11 \xrightarrow{} 22 \xrightarrow{} 21$ &
                      - &
                      - &
                      - &
                      $21 \xrightarrow{} 42 \xrightarrow{} 12$ \\
                      
\bottomrule

\end{tabular}
}
\caption{\emph{Model details.} Each row represents a different model hyperparameter and each column a different model, where ``HyperSeg-$<$size$>$ (dataset)'' is the model's template name (see Sec.~\ref{sec:Experimental results}) and the dataset on which it was trained on. ``-'' denotes that the decoder level corresponding to the specific hyperparameter was omitted.}
\label{tab:model_details}
\end{table*}

\section{Additional ablation studies}
\minisection{Ablation study on PASCAL VOC}
We tested multiple variants of our method, to show the effects of employing spatially varying convolutions and to evaluate the contribution of the positional encoding. We describe these variants using the following terminology: $1 \times 1$ denotes evaluating the entire image as a single patch. That is, we only generate a single set of weights per input image. For this variant, we completely remove network $h$. $16 \times 16$ is the original number of patches used for reporting our results on PASCAL VOC in Tab.~\ref{tab:pascal_voc_val_quant}. 

Our ablation results are reported in Tab.~\ref{tab:pascal_voc_val_ablation}. Evidently, the larger the grid size, the better our accuracy. The contribution of the positional encoding is significant, given that the gap between the two best previous methods is 0.1\%, as can be seen in Tab.~\ref{tab:pascal_voc_val_quant}. As evident from the reported FPS column, the $1 \times 1$ variant is the fastest because it does not require unoptimized operations used for the DPWConv and because the network $h$ is absent. 

\begin{table}[t!]
\centering{
\begin{tabular}{c c c c }
  \toprule
  Grid size & Positional encoding & mIoU (\%) & FPS \\ [0.5ex] 
  \hline
  $1 \times 1$   & \textcolor{red}{\xmark} & 77.56 & \textbf{46.8} \\
  $4 \times 4$   & \textcolor{red}{\xmark} & 78.92 & 22.4 \\
  $8 \times 8$   & \textcolor{red}{\xmark} & 80.23 & 26.9 \\
  $16 \times 16$ & \textcolor{red}{\xmark} & 80.33 & 28.2 \\
  $16 \times 16$ & \textcolor{darkgreen}{\checkmark} & \textbf{80.61} & 26.8 \\
  \bottomrule
\end{tabular}
\vspace{1mm}
\caption{\emph{Ablation study on PASCAL VOC 2012, val. set~\cite{everingham2010pascal}.}
Each row represents a different model, trained with the specified grid size, with or without positional encoding.}
\vspace{-3mm}
}
\label{tab:pascal_voc_val_ablation}
\end{table}

\minisection{Ablation study on Cityscapes}
We test the effect of varying $g_{w_i},i\in[1,5]$ in our HyperSeg-M model on resolutions of $1536 \times 768$ and report results in Tab.~\ref{tab:cityscapes_val_groups_ablation}. We fix the ratio between the groups relative to $|\theta^{m_i}|$ while maintaining multiples of 2. We start by setting $g_{w_1}=1$ for the test in the first row and then double the group number in each subsequent test. The number of parameters and flops of $w_i$ decreases as the number of groups increases, according to Eq.~\ref{eq:params} and Eq.~\ref{eq:flops}. Surprisingly, the experiment in the middle row provides the best GFLOPs / accuracy trade-off, achieving the best accuracy with fewer GFLOPs and parameters than the first two tests.

\begin{table}[t!]
\centering{
\begin{tabular}{c c c c c }
  \toprule
  \multirow{2}{*}{Groups} & mIoU & \multirow{2}{*}{GFLOPs} & $|\theta^w|$ & Params   \\ [0.5ex] 
  & (\%) & & (M) & (M) \\
  \hline
  1, 2, 4, 4, 8       & 77.1          & 18.0 & 1.2 & 11.1 \\
  2, 4, 8, 8, 16      & 77.5          & 17.3 & 0.6 & 10.4 \\
  4, 8, 16, 16, 32    & \textbf{78.0} & 16.9 & 0.3 & 10.1 \\
  8, 16, 32, 32, 64   & 77.8          & 16.7 & 0.1 & 10.0 \\
  16, 32, 64, 64, 128 & 77.2          & \textbf{16.6} & \textbf{0.1} & \textbf{9.9} \\
  \bottomrule
\end{tabular}
}
\caption{\emph{Ablation study on Cityscapes, val. set~\cite{cordts2016cityscapes}. The {\em Groups} column represents: $g_{w_1},\dots,g_{w_n}$.}
\vspace{-2mm}}
\label{tab:cityscapes_val_groups_ablation}
\end{table}

\section{Open source repositories}
\label{sec:Open source repositories}
The open source repositories used in Tables~\ref{tab:pascal_voc_val_quant}, \ref{tab:cityscapes_quant}, and \ref{tab:camvid_quant}, to report information about previous methods, which was not available otherwise, are listed in Tab.~\ref{tab:open_source}. The protocol for computing the FPS, GFLOPs, and trainable parameters, is described in Sec.~\ref{sec:Experimental results}.

\begin{table*}[t!]
\centering{
\begin{tabular}{@{}ll@{}}
\toprule
Method & URL \\ \midrule
Auto-DeepLab-L~\cite{liu2019auto}       & \url{https://github.com/MenghaoGuo/AutoDeeplab}   \\
DeepLabV3~\cite{chen2017rethinking}     & \url{https://github.com/pytorch/vision}           \\
ERFNet~\cite{romera2017erfnet}          & \url{https://github.com/Eromera/erfnet_pytorch}   \\
ESPNetV2~\cite{mehta2018espnet}         & \url{https://github.com/sacmehta/ESPNetv2}        \\
SwiftNetRN-18~\cite{orsic2019defense}   & \url{https://github.com/orsic/swiftnet}           \\
BiSeNetV1~\cite{yu2018bisenet}          & \url{https://github.com/CoinCheung/BiSeNet}       \\
BiSeNetV2~\cite{yu2020bisenet}          & \url{https://github.com/CoinCheung/BiSeNet}       \\
\bottomrule
\end{tabular}
}
\caption{\emph{List of open source repositories used for comparing to previous methods.} }
\label{tab:open_source}
\end{table*}

\section{Convolution and batch normalization fusion}
\label{sec:Batch normalization fusion}
Following others~\cite{orsic2019defense}, we fuse the convolution and batch normalization operations in the inference stage for improving the runtime performance. For regular convolutions, the batch normalization operation is fused with its prior convolution. The batch normalization operations following DPWConv are fused with the corresponding weight mapping layer. We next provide more details on these steps.

The batch normalization operation can be described in the form of matrix multiplication, $\theta^{BN} \cdot x + b^{BN}$, 
for which $\theta^{BN}_{i,i}=\frac{\gamma_i}{\sqrt{\sigma^2_i+\epsilon}}$ on the diagonal and zero everywhere else, 
and $b^{BN}_i=\beta_i - \gamma_i \frac{\mu_i}{\sqrt{\sigma^2_i+\epsilon}}$ for $i \in [0,C)$, where $C$ is the number of feature channels, $\gamma$ is the scaling factor, $\beta$ is the shifting factor, $\mu$ and $\sigma$ are the mean and standard deviation, respectively, and $\epsilon$ is a scalar constant used for numeric stabilization.

The convolution operation can also be written as a matrix multiplication by reshaping its weights, $\theta^*$, and input feature map, $x$, such that $\theta^* \in \mathbb{R}^{C \times C_{in} \times k^2}$ and $\tilde{x} \in \mathbb{R}^{C_{in} \times k^2}$, where $\tilde{x}_{i,j}$ is the $k \times k$ neighborhood in location $(i,j)$ of the original feature, $x$. The combined convolution and batch normalization operations can then be written as:
\begin{equation}
    O_{i,j} = \theta^{BN} \cdot (\theta^{*} \cdot \tilde{x}_{i,j} + b^*) + b^{BN}.
\end{equation}
The fused convolution operation will then have the weights $\tilde{\theta}^* = \theta^{BN} \cdot \theta^*$ and bias 
$\tilde{b}^* = \theta^{BN} \cdot b^{*} + b_{BN}$.

Similarly, the DPWConv operation on a patch location $(m,n)$ followed by batch normalization is:
\begin{equation}
    O_{i,j} = \theta^{BN} \cdot \left[(\theta^w \cdot \phi_{m,n}) \cdot \tilde{x}_{i,j} + b^w \right] + b^{BN},
\end{equation}
where $\theta^w$ are the weights of the weight mapping layer and $\phi_{m,n}$ is the signal corresponding to the patch in location $(m,n)$.
The batch normalization operation can then be fused into the weight mapping layer using the adjusted weights, 
$\tilde{\theta}^w = \theta^{BN} \cdot \theta^{w}$, and bias, $\tilde{b}^w = \theta^{BN} \cdot b^w + b^{BN}$.

\section{Additional qualitative results}
\label{sec:Additional qualitative results}

\begin{figure*}[t]
\centering
\includegraphics[clip,trim=0mm 0mm 0mm 0mm, width=1.0\textwidth]{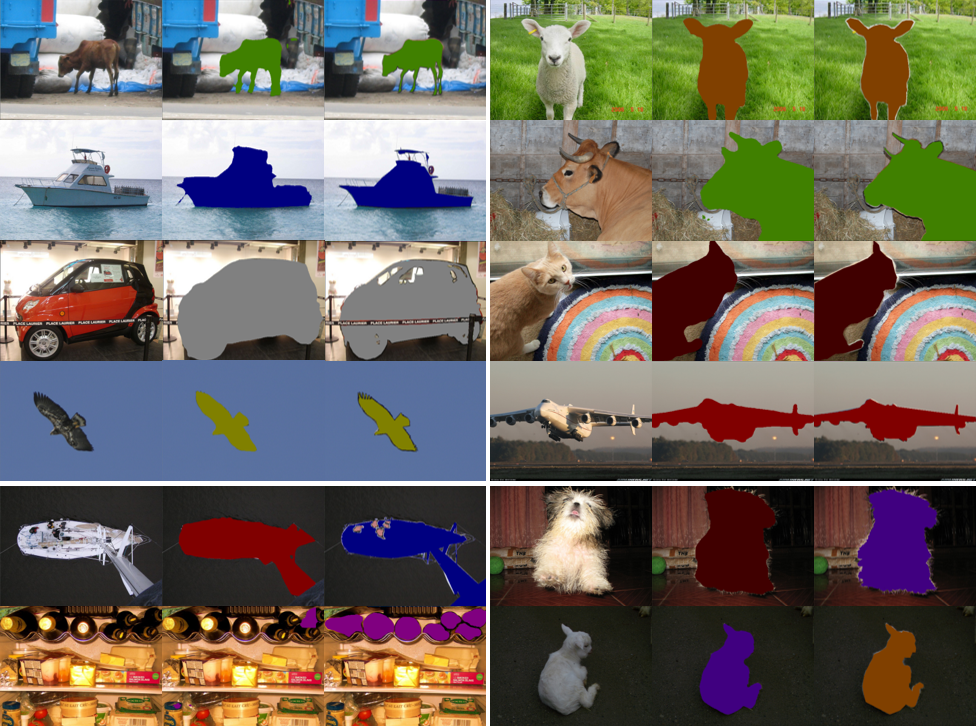}\vspace{0mm}
\caption{
\emph{Qualitative results on PASCAL VOC 2012 validation set~\cite{everingham2010pascal}.} First and 4th columns: input image, 2nd and 5th columns: our predictions, and 3rd and final column: ground truth. The first four rows demonstrate how our model performs on different classes. The last two rows present failure cases of our model.
}
\label{fig:pascal_voc_val_qualitative}
\end{figure*}

Fig.~\ref{fig:pascal_voc_val_qualitative} provides qualitative results on PASCAL VOC 2012 val. set~\cite{everingham2010pascal}. In the first four rows we have specifically chosen samples with different classes to best demonstrate the performance of our model. The last two rows offer failure cases, top left: boat classified as a chair; bottom left: the model failed to detect the bottles from a top view; top right: dog classified as a cat; and bottom right: sheep classified as a dog.

\begin{figure*}[t]
\centering
\includegraphics[clip,trim=0mm 0mm 0mm 0mm, width=0.8\textwidth]{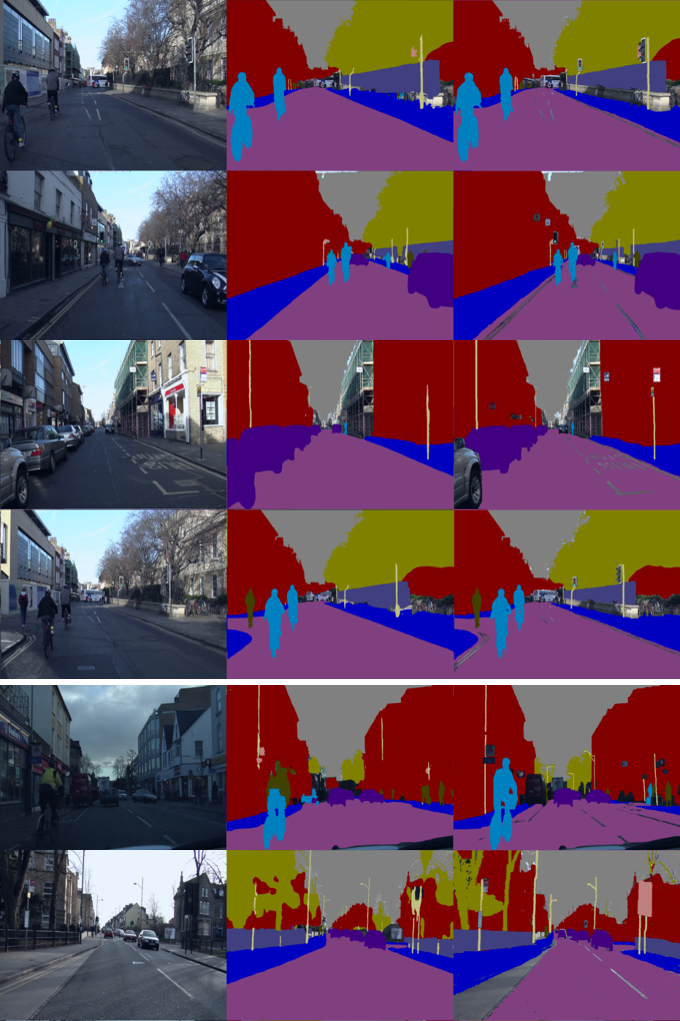}\vspace{0mm}
\caption{
\emph{Qualitative results on CamVid test set~\cite{brostow2009semantic}.} The columns represent: input (left), prediction (center), and ground truth (right). The first four rows provide samples from different scenes, and the last two rows demonstrate failure cases.
}
\label{fig:camvid_test_qualitative}
\end{figure*}

Fig~\ref{fig:camvid_test_qualitative} shows qualitative results of our HyperSeg-L model on the CamVid dataset test set~\cite{brostow2009semantic}. The first four rows display predictions on different scenes and the last two rows demonstrate failures of our model: in the first row a bicyclist is partly segmented as a pedestrian, and in the last row our model fails to detect a sign.

\end{document}